\documentclass{article}

\usepackage{docmute}

\usepackage{arxiv}

\usepackage[utf8]{inputenc}
\usepackage[T1]{fontenc}

\usepackage{amsmath}
\usepackage{amssymb}
\usepackage{amsfonts}
\usepackage{amsthm}
\usepackage{bm}
\usepackage{comment}
\usepackage{fancybox}
\usepackage{framed}
\usepackage{color}
\usepackage[dvipdfmx]{graphicx}
\usepackage{multicol}
\usepackage{multirow}
\usepackage{pdflscape}
\usepackage{verbatim}
\usepackage{subfigure}
\usepackage{cite}
\usepackage{hyperref}
\usepackage{url}

\hypersetup{
    colorlinks = true,
    urlcolor = blue,
    linkcolor = red,
    citecolor = green
}

\usepackage{algorithm}
\usepackage[noend]{algpseudocode}
\usepackage{algorithmicx}

\expandafter\def\expandafter\UrlBreaks\expandafter{\UrlBreaks
  \do\a\do\b\do\c\do\d\do\e\do\f\do\g\do\h\do\i\do\j%
  \do\k\do\l\do\m\do\n\do\o\do\p\do\q\do\r\do\s\do\t%
  \do\u\do\v\do\w\do\x\do\y\do\z\do\A\do\B\do\C\do\D%
  \do\E\do\F\do\G\do\H\do\I\do\J\do\K\do\L\do\M\do\N%
  \do\O\do\P\do\Q\do\R\do\S\do\T\do\U\do\V\do\W\do\X%
  \do\Y\do\Z}

\algrenewcommand\algorithmicindent{1.0em}

\algnewcommand\algorithmicforeach{\textbf{for each}}
\algdef{S}[FOR]{ForEach}[1]{\algorithmicforeach\ #1\ \algorithmicdo}

\algnewcommand\AlgAnd{\textbf{and} }
\algnewcommand\AlgOr{\textbf{or} }

\algrenewcommand\textproc{}

\algnewcommand{\Initialize}[1]{
	\State \textbf{Initialize:}
 	\State \hspace*{\algorithmicindent}\parbox[t]{0.8\linewidth}{\raggedright #1}}

\title{A Sequential Concept Drift Detection Method for On-Device Learning on Low-End Edge Devices}

\author{
  Takeya Yamada\\
  Keio University\\
  3-14-1 Hiyoshi, Kohoku-ku, Yokohama, Japan\\
  \texttt{takeya@arc.ics.keio.ac.jp}\\
  \And
  Hiroki Matsutani \\
  Keio University\\
  3-14-1 Hiyoshi, Kohoku-ku, Yokohama, Japan\\
  \texttt{matutani@arc.ics.keio.ac.jp} \\
}

\begin{document}

\maketitle

\begin{abstract}
A practical issue of edge AI systems is that data distributions of trained dataset and deployed environment may differ due to noise and environmental changes over time.
Such a phenomenon is known as a concept drift, and this gap degrades the performance of edge AI systems and may introduce system failures. 
To address this gap, retraining of neural network models triggered by concept drift detection is a practical approach.
However, since available compute resources are strictly limited in edge devices, in this paper we propose a fully sequential concept drift detection method in cooperation with an on-device sequential learning technique of neural networks.
In this case, both the neural network retraining and the proposed concept drift detection are done only by sequential computation to reduce computation cost and memory utilization.
Evaluation results of the proposed approach shows that while the accuracy is decreased by 3.8\%-4.3\% compared to existing batch-based detection methods, it decreases the memory size by 88.9\%-96.4\% and the execution time by 1.3\%-83.8\%. As a result, the combination of the neural network retraining and the proposed concept drift detection method is demonstrated on Raspberry Pi Pico that has 264kB memory.
\end{abstract}


\section{Introduction} \label{sec:intro}
With the rapid spread of AI (Artificial Intelligence) and IoT (Internet-of-Things) technologies, the number of IoT devices connected to the Internet continues to grow significantly.
In cloud-based AI systems, IoT devices typically collect data at deployed edge environments and send the data to datacenters via the Internet.
In this case, IoT devices focus on the data collection, and cloud servers are in charge of big data analysis and sophisticated machine learning tasks using plenty of compute resources.
In addition, edge intelligence \cite{Zhou19} in which some machine learning tasks such as prediction are performed in the edge side is also becoming popular since performance and efficiency of edge devices have been improved significantly.
Although conventional edge AI systems focus on prediction tasks, recently an on-device learning approach of neural networks is proposed for resource-limited IoT devices \cite{Matutani}.

However, there are some limitations on such on-device learning approaches.
First, there is a limitation on computation power and cost. Since they are often battery-powered, low-power consumption is required.
In addition, deployed environments around the edge devices may change over time. That is, data distribution observed by the edge devices may shift as time goes by.
For example, data distributions of trained dataset and deployed environment may differ due to noise and environmental changes.
This gap degrades the performance of edge AI systems and may introduce system failures. To address this gap, a concept drift detection is a well-known approach \cite{OASW}.

Since edge devices are resource-limited, in this paper we propose a
lightweight concept drift detection method for resource-limited edge
devices.
The contributions of this paper are as follows.
\begin{itemize}
\item We propose a fully sequential concept drift detection
  method to be combined with the on-device sequential learning approach of
  neural networks.
\item Since both the neural network retraining and the proposed concept
  drift detection are done only by sequential computation, we
  demonstrate that the combined approach is implemented on Raspberry Pi
  Pico that has 264kB memory.
  \end{itemize}
The proposed approach is compared to existing concept drift detection methods in terms of accuracy using practical datasets.
They are also evaluated in terms of execution time and memory utilization on edge devices.

The rest of this paper is organized as follows. Section \ref{sec:related} overviews concept drift detection methods.
Section \ref{sec:design} explains the proposed detection method.
Section \ref{sec:env} describes the experimental setup including the datasets, counterparts, and platforms.
Section \ref{sec:eval} shows the evaluation results in terms of the accuracy, execution time, and memory utilization.
Section \ref{sec:conc} concludes this paper.

\section{Background and Related Work}\label{sec:related}
\subsection{Concept Drift Types}\label{sec:concepttype}
\begin{figure}[h]
    \centering
    \includegraphics[width=87mm]{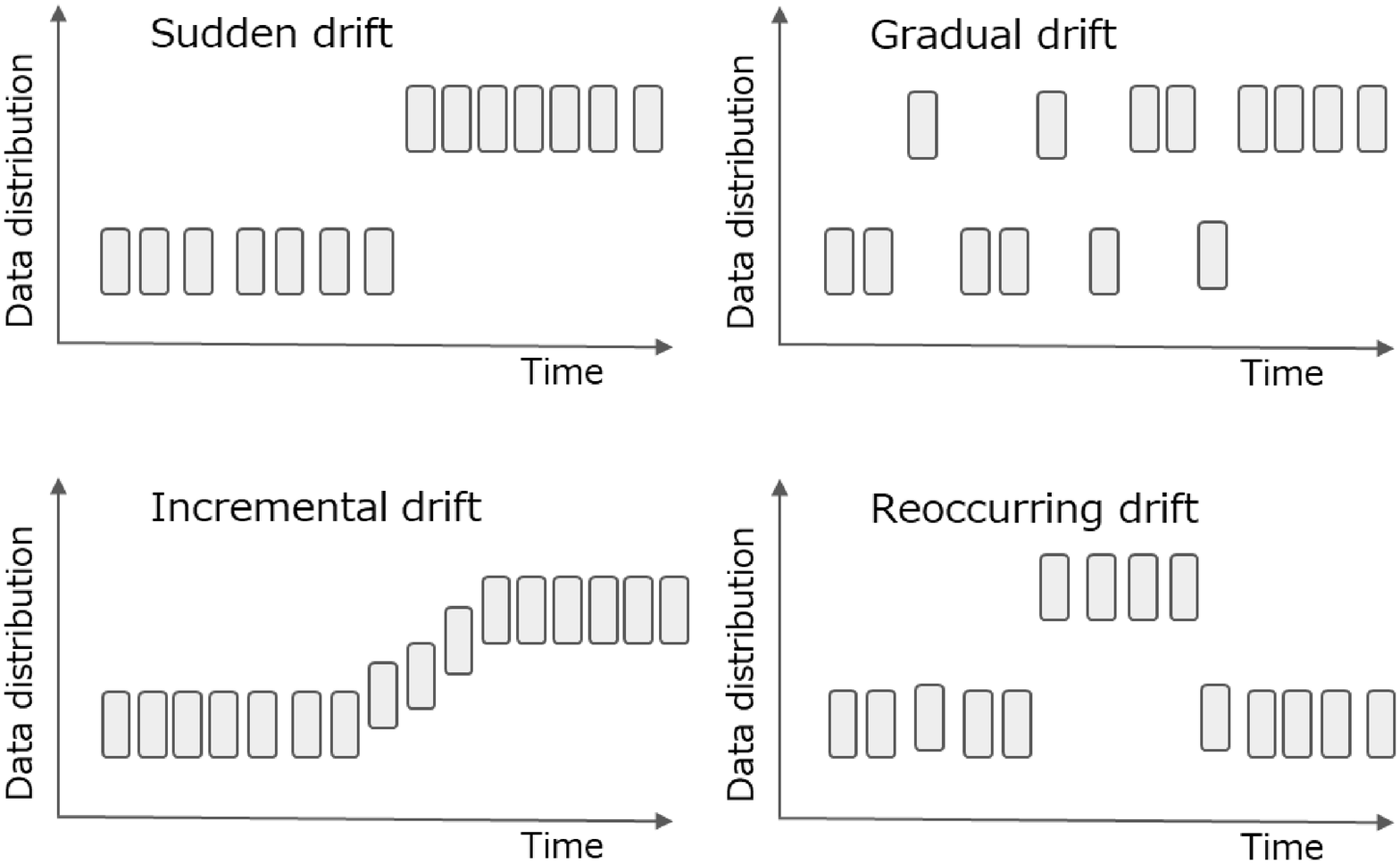}
    \caption{Four concept drift types \cite{Survey}}
    \label{fig:concepttype}
\end{figure}

A concept drift \cite{Survey} is known as a phenomenon where statistical properties of target data change over time.
It is sometimes caused by changes on hidden variables which cannot be observed directly.
There are various types of concept drifts, and representative ones \cite{Survey} are illustrated in Figure \ref{fig:concepttype}.
In the figure, the vertical and horizontal axes represent data distribution and elapsed time, respectively.
The sudden drift is a concept drift in which a data distribution changes suddenly.
In the sudden drift, an old data distribution before the concept drift does not appear after the concept drift.
The gradual drift is a concept drift in which an old data distribution is gradually replaced with a new data distribution. Both the old and new distributions appear during the concept drift.
In the incremental drift, the data distribution is incrementally shifted from an old distribution to a new distribution during the concept drift.
In the reoccurring drift, after the data distribution has been changed to a new one, the old data distribution reoccurs.

\subsection{Concept Drift Countermeasures}\label{adapt_mehtod}
There are various approaches to address the concept drifts, and they can be classified into active approaches and passive approaches \cite{NOVELOSELM}.
In this paper, a machine learning model that solves classification or regression tasks is called a ``discriminative model'', and a model that detects concept drifts is called a ``detection model''. An example of their relationship is illustrated in Figure \ref{fig:overview}.

\subsubsection{Passive Approach}
In the passive approach, a discriminative model is retrained whenever a new data arrives.
Since the discriminative model can be always trained with the latest data, its accuracy tends to be high.
It does not use any detection model.
However, it requires computation resource and memory to retrain a discriminative model. This may limit its application to resource-limited edge devices. 
To enable retraining of neural networks on resource-limited edge devices, OS-ELM \cite{OSELM} is used as an online sequential learning algorithm of 3-layer neural networks in \cite{OSUAD}.
Since it sequentially updates weight parameters of neural networks when new training samples come, the memory utilization is quite small compared to batch training algorithms.
ONLAD \cite{OSUAD} is classified as a passive approach.
Specifically, OS-ELM is combined with a forgetting mechanism to forget old collected data and follow a concept changes quickly. The training batch size is fixed to one so that pseudo inverse operation of matrixes can be eliminated.

\subsubsection{Active Approach}\label{sssec:active}
In the active approach, a machine learning model is retrained only when a concept drift is detected. It thus requires a detection model in addition to a discriminative model.
Since it may be difficult to address all the concept drift types introduced in Section \ref{sec:concepttype} at the same time for any applications, existing active approaches often focus on some specific concept drift types \cite{Survey2}.
There are several detection models and they can be broadly classified into two methods below and their ensemble \cite{Survey}.

The first detection method is an error-rate based drift detection method.
This method monitors prediction errors of a discriminative model using labeled teacher data, and it detects a concept drift when the error-rate exceeds a threshold value.
DDM \cite{DDM} and ADWIN \cite{ADWIN} are the error-rate based drift detection methods.
DDM (Drift Detection Method) uses two threshold levels: warning level and drift level. When an error-rate reaches the warning level, it starts a retraining of a discriminative model. When the error-rate reaches the drift level, the retrained discriminative model replaces the old model.
The number of samples required to judge concept drifts, which is called window size, is fixed at DDM.
In ADWIN (Adaptive Windowing), the window size is adaptively adjusted based on test statistics.
Since these approaches need a labeled teacher dataset to detect a concept drift, they are not suited to resource-limited edge devices with a limited memory capacity.

The second detection method is a distribution-based drift detection.
Quant Tree \cite{QuantTree} and SPLL \cite{SPLL} are the distribution-based drift detection methods.
Quant Tree detects concept drifts by using a histogram.
Although the size of histogram increases as the number of features (the number of dimensions) increases in typical histogram-based detection methods, it can be fixed in Quant Tree.
Also, the test statistics to detect concept drifts does not depend on training and test datasets.
SPLL detects concept drifts by using semi-parametric log-likelihood.
Input data samples are clustered by using k-means method, and then the resultant clusters are modeled assuming GMM (Gaussian Mixture Model) to detect concept drifts.
The distribution-based drift detection methods, such as Quant Tree and SPLL, often process a batch of data samples to detect concept drifts. Thus, they are also not suited to resource-limited edge devices with a limited memory capacity.
In this paper, we will propose a concept drift detection method of neural networks both of which require only sequential computation to reduce computation cost and memory utilization.

\section{Proposed Detection Method}\label{sec:design}
\begin{figure}[h]
	\centering
	\includegraphics[width=70mm]{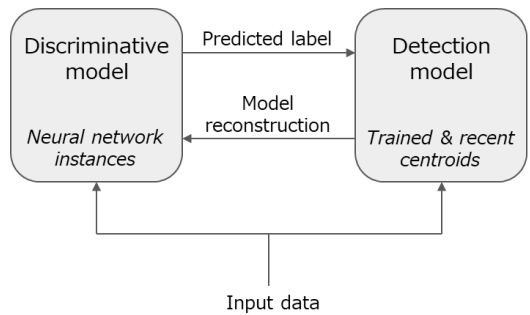}
\caption{Overview of proposed detection method}
\label{fig:overview}
\end{figure}

Figure \ref{fig:overview} illustrates an overview of the proposed concept drift detection method.
Specifically, the proposed lightweight detection method is combined with the on-device sequential learning approach of neural networks \cite{Matutani} as a discriminative model.
In this section, the discriminative model assumed in this paper is briefly illustrated first. Then the proposed detection method is explained.

\subsection{Discriminative Model}\label{ssec:model} 
Assume data can be classified into one of multiple labels.
In the discriminative model, the same number of OS-ELM based neural networks (called ``instances'') as the number of labels in the training dataset are used.
For each label in the training dataset, a discriminative model instance is trained with the data belonging to the label.
Each discriminative model instance forms an autoencoder \cite{Hinton06} for unsupervised anomaly detection. That is, the numbers of input and output layer nodes of the discriminative model instances are the same, and each instance is trained so that its output can reconstruct a given input data with a smaller number of hidden nodes.

In the test phase, a reconstruction error (anomaly score) is calculated by comparing the input and output data in each model instance, and the smallest anomaly score among all the instances is used as the final prediction result (see lines 6 and 7 in Algorithm \ref{alg:main}). For the sequential training, a single model instance that outputs the smallest anomaly score (i.e., the ``closest'' instance) trains the input data sequentially.

\subsection{Concept Drift Detection} 
At first, the proposed concept drift detection method calculates a centroid of trained data for each label. It records the same number of trained centroids as the number of labels. Then it sequentially updates the centroid with recent test data for each label whenever it predicts. It maintains recent centroids in addition to the trained centroids.
A drift rate is calculated based on a sum of the distance between the trained centroid and corresponding recent centroid for each label (see line 14 in Algorithm \ref{alg:main}). Then a concept drift is detected when the drift rate exceeds a pre-determined threshold value.

\begin{figure}[h]
	\centering
	\begin{minipage}[b]{0.245\linewidth}
	  \includegraphics[width=40mm]{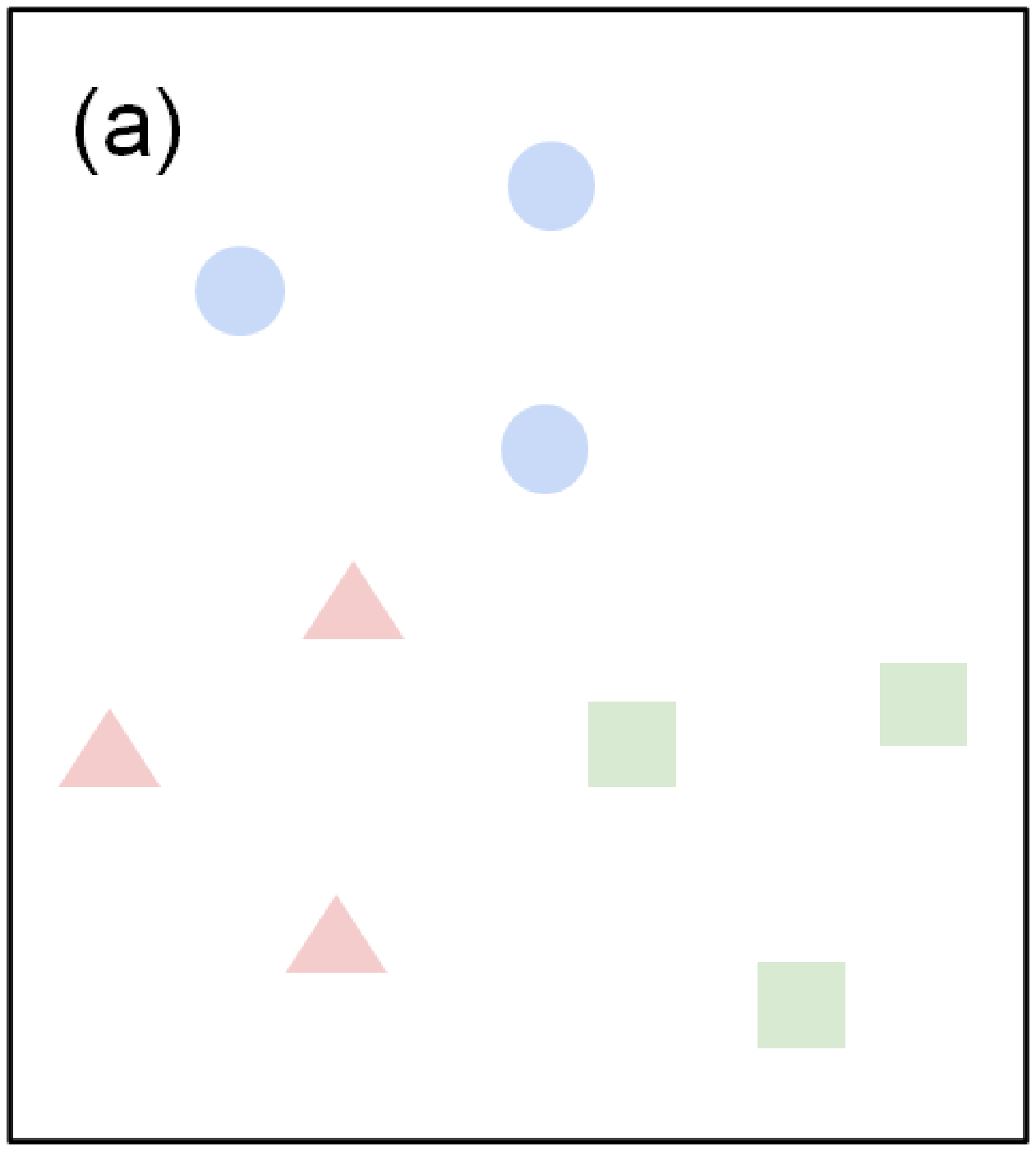}
	\end{minipage}		
	\begin{minipage}[b]{0.245\linewidth}
	  \includegraphics[width=40mm]{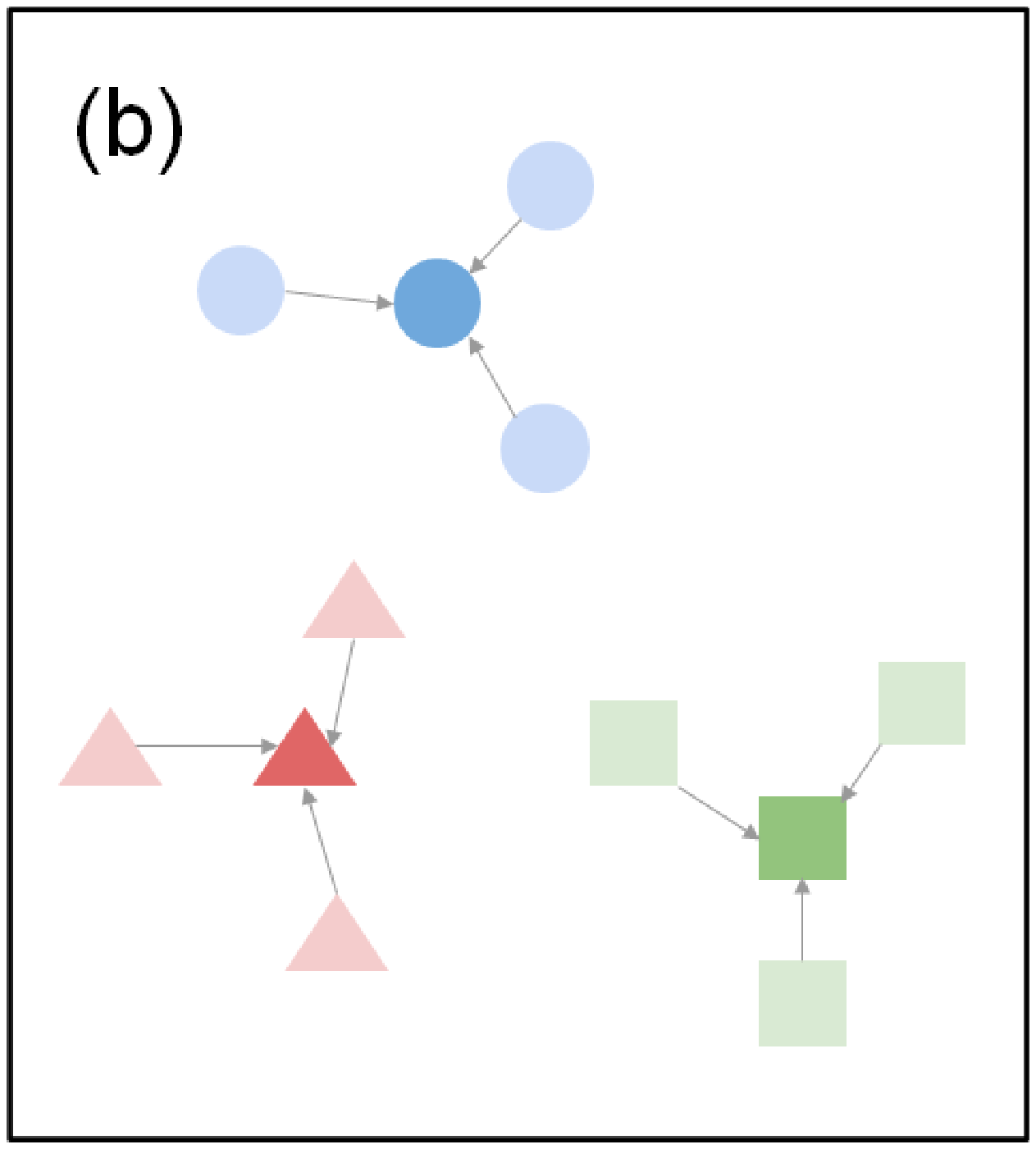}
	\end{minipage}
	\centering
	\begin{minipage}[b]{0.245\linewidth}
	  \includegraphics[width=40mm]{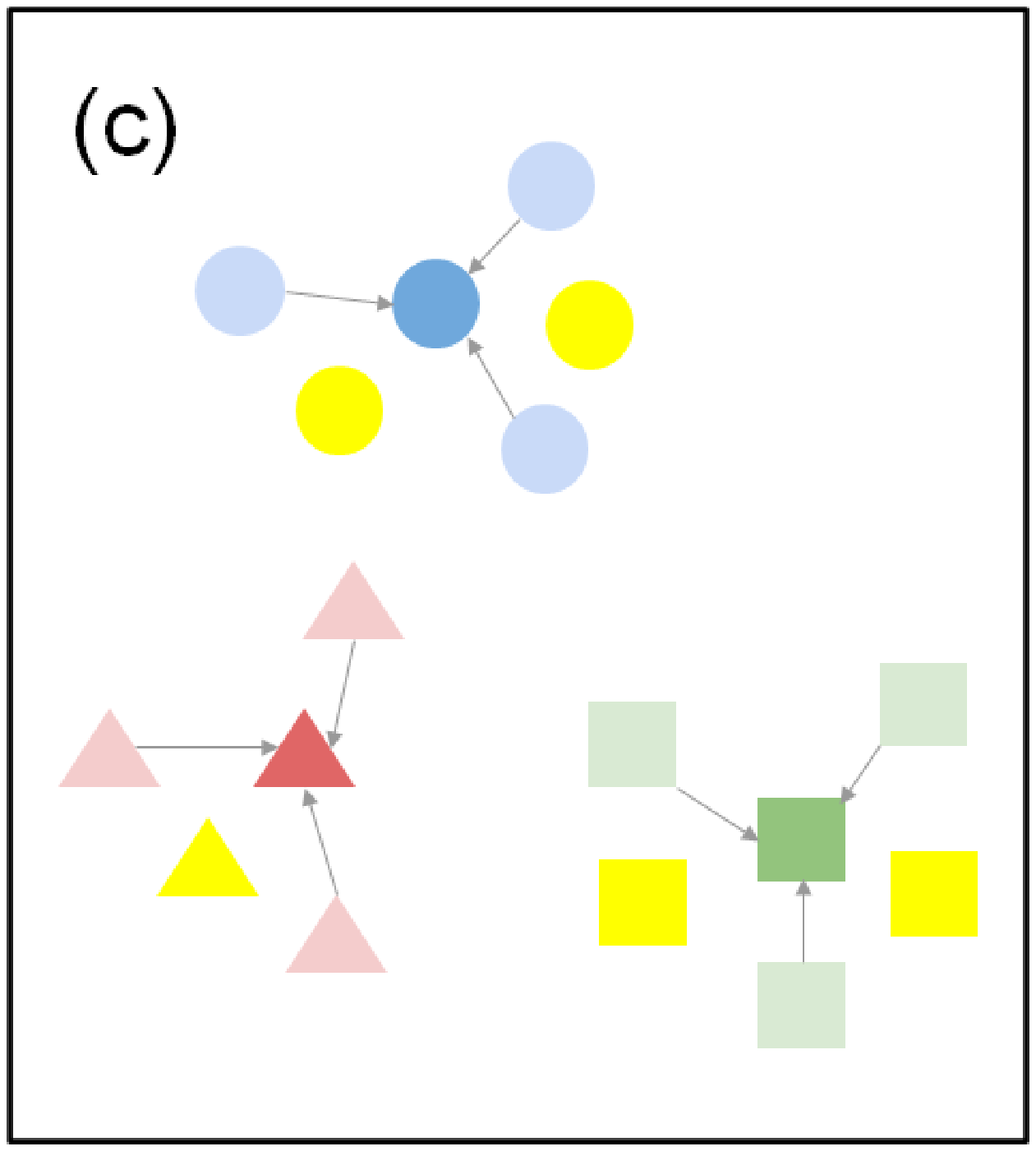}
	\end{minipage}		
	\begin{minipage}[b]{0.245\linewidth}
	  \includegraphics[width=40mm]{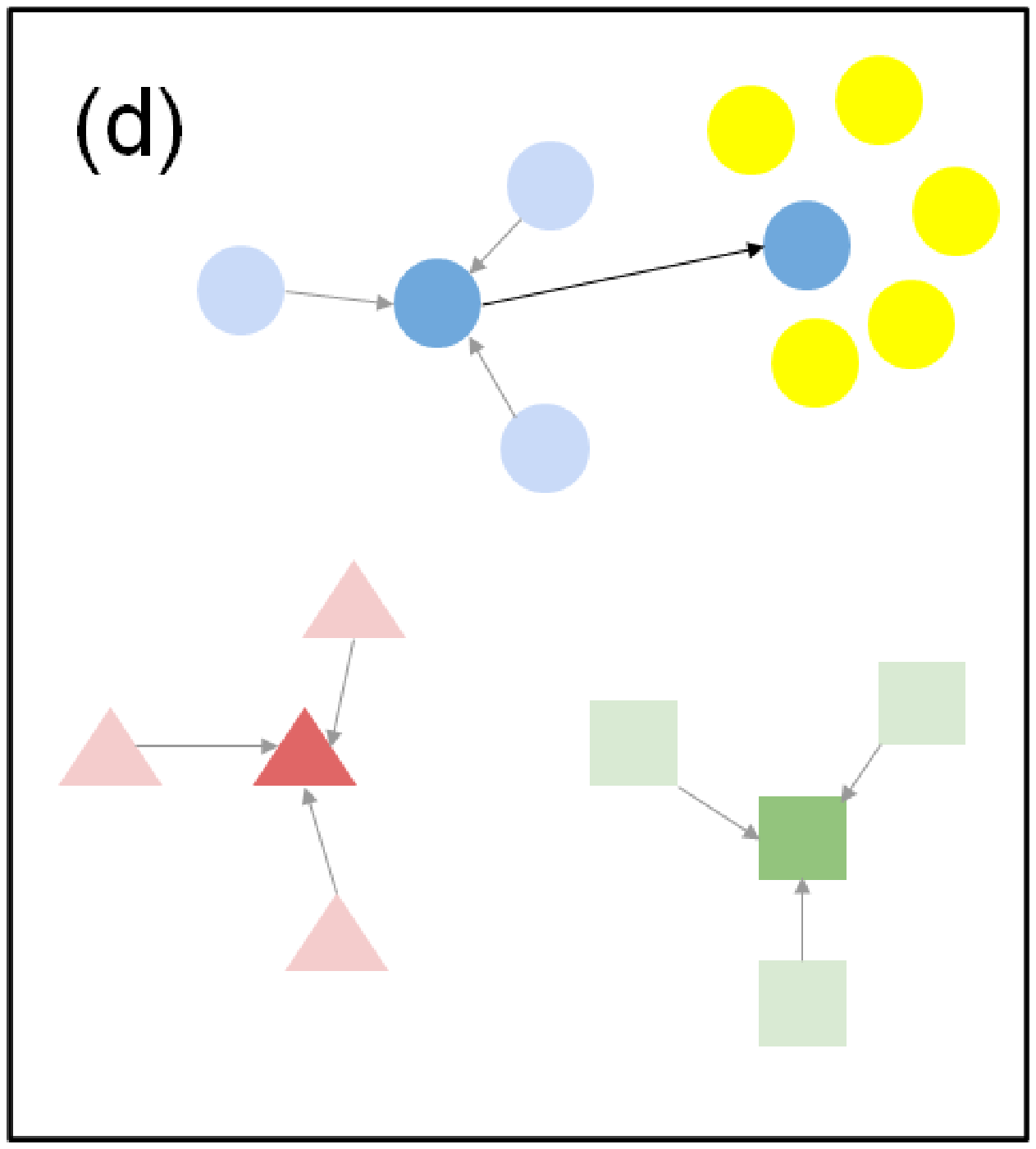}
	\end{minipage}		
	\\
\caption{Overview of proposed detection algorithm: (a) initial samples, (b) trained centroids, (c) test centroids before concept drift, and (d) those after concept drift}
\label{fig:state}
\end{figure}

The proposed method is illustrated below.
In the initial training phase, a discriminative model is trained with initial samples.
Assume there are initial samples which are labeled as one of three different colors as shown in Figure \ref{fig:state}(a).
In the case of unsupervised learning, it is assumed that these initial samples can be labeled with a clustering algorithm such as k-means.
A centroid of the initial samples is calculated for each label, as shown in Figure \ref{fig:state}(b). In this figure, the centroids are represented as deeper-colored points. They are referred as ``trained centroids''. The proposed method thus records the same number of trained centroids as the number of labels in the initial training phase.

In the prediction phase, the discriminative model predicts a label for each test sample. The centroids are sequentially updated based on each test sample and its predicted label. They are referred as ``test centroids''. When the centroids are calculated, it is possible to assign a higher weight to a newer sample (a lower weight to an older sample) so that they can represent ``recent'' test centroids.
Assume a new test sample comes and it is labeled as ``blue'' by the discriminative model. The recent test centroid of ``blue'' label is then sequentially updated. The data distribution is relatively stable before a concept drift occurs. We can thus expect that distances between the trained centroids and the recent test centroids are small as shown in Figure \ref{fig:state}(c).

Next, let us illustrate a case when a concept drift happens.
Assume the data distribution is changed and new test samples appear as shown in yellow circles in Figure \ref{fig:state}(d).
If these new test samples are labeled as ``blue'' by the discriminative model, the recent test centroid of ``blue'' label is moved to near the yellow circles in Figure \ref{fig:state}(d). The distance between the test centroid and the trained centroid increases due to the new data distribution.
A drift rate is calculated based on a sum of these distances, and a concept drift is detected when the drift rate exceeds a pre-determined threshold value.
Please note that in the proposed method, the distances can be sequentially updated when it predicts. It thus requires much less memory than batch-based concept drift detection methods introduced in Section \ref{sssec:active}.

Algorithm \ref{alg:main} shows the proposed detection method mentioned above.
The inputs to the proposed algorithm are a discriminative model, a test dataset, the number of class labels, the number of dimensions of the data, trained centroids, the number of samples in each label, a window size, and error/drift threshold values.
Variable $drift$ indicates whether a concept drift is occurring now or not. In line 2, it is initialized to False.
Variable $check$ indicates whether a concept drift needs to be checked or not. In line 3, it is initialized to False.
In line 6, assuming the discriminative model with multiple instances introduced in Section \ref{ssec:model}, a class label predicted by the model is set to variable $c$.
In line 7, an anomaly score predicted by the discriminative model is set to variable $error$.
In lines 8 and 9, if the anomaly score exceeds a given threshold value $\theta_{error}$, $check$ is set to True so that a concept drift will be checked. $\theta_{error}$ is a tuning parameter.
In line 10, a window counter $win$ is initialized to 0.

If $check$ is True and the window counter is less than a pre-determined window size $W$, recent test centroid of the predicted label $c$ is sequentially updated based on an incoming $data$. In line 12, $cor[c]$ is the test centroid of label $c$. Then a sum of the distance between the trained centroid and the recent test centroid for each label is updated. In line 14, $dist$ is the sum of the distances.
If the window counter $win$ reaches a given window size $W$, a concept drift is checked by comparing the distance $dist$ and a threshold value $\theta_{drift}$ in line 17. The threshold $\theta_{drift}$ can be determined as explained in Section \ref{ssec:thr}.
If the distance exceeds the threshold, $drift$ is set to True in line 18.
In this case, the discriminative model is retrained in line 21.
Reconstruct\_Model() function is described in Section \ref{sec:detaiL_prop}.

\begin{algorithm}[h]
	\caption{Concept drift detection}
	\label{alg:main}
	\begin{algorithmic}[1]
		\Require discriminative model $model$, test dataset $testdata$, number of classes $C$, number of dimensions $D$, recent coordinates $cor$, trained coordinates $train\_cor$, numbers of samples in classes $num$, window size $W$, error threshold $\theta_{error}$, drift threshold $\theta_{drift}$
		\Function{Main}{}
			\State $drift \gets \mathrm{False}$
			\State $check \gets \mathrm{False}$
			\For{$data \in testdata$}:
				\If{$drift = \mathrm{False} \And check = \mathrm{False}$}
					\State $c \gets \mathrm{argmin}_{i \in C}model[i].\mathrm{predict}(data)$
					\State $error \gets model[c].\mathrm{predict}(data)$
					\If{$error \geq \theta_{error}$}
							\State $check \gets \mathrm{True}$
							\State $win \gets 0 $
					\EndIf
				\EndIf
				\If{$check = \mathrm{True} \And win < W$}
					\State $cor[c] \gets \dfrac{cor[c] \times num[c] + data}{num[c]+1}$
					\State $num[c] \gets num[c] + 1$
					\State $dist \gets \Sigma_{i \in C}\Sigma_{j \in D}{|cor[i][j] - train\_cor[i][j]|}$
					\State $win \gets win + 1$
					\If{$win = W$}
						\If{$dist \geq \theta_{drift}$}
							\State $drift \gets \mathrm{True}$
						\EndIf
						\State $check \gets \mathrm{False}$
					\EndIf
				\EndIf
				\If{$drift = \mathrm{True}$}
					\State $drift \gets$ Reconstruct\_Model($data$)
				\EndIf
				
			\EndFor
		\EndFunction
	\end{algorithmic}
\end{algorithm}

\subsection{Model Reconstruction}\label{sec:detaiL_prop}
The discriminative model reconstruction is divided into four parts.
The model reconstruction flow is shown in Algorithm \ref{alg:reconst}.
In the first part, an initial coordinate is selected for each label by Init\_Coord() function in line 4.
In Init\_Coord() function, given the number of labels $C$, $C$ initial samples are selected as initial coordinates of $C$ labels.
Specifically, these initial coordinates are selected so that a sum of all the distances between these coordinates is maximized.
This part is inspired from an idea of k-means++ algorithm \cite{Arthur07} that spreads out initial cluster centroids for a better clustering.
The detail implementation of Init\_Coord() function is shown in Algorithm \ref{alg:initcoord}.

In the second part, centroids of $C$ labels are updated based on an incoming data by Update\_Coord() function in line 6.
Specifically, a label that minimizes the distance between the incoming data and centroid of the label is selected. Based on the selected label, its centroid is then sequentially updated.
This part is very similar to a sequential k-means algorithm.
The detail implementation of Update\_Coord() function is shown in Algorithm \ref{alg:updatecoord}.
Please note that since there is a possibility that initial coordinates selected by Init\_Coord() are outliers, the centroids are further refined by Update\_Coord() function using more samples.

In the third part, the discriminative model is retrained.
In lines 8 and 9, a label that minimizes the distance between the incoming data and centroid of the label is selected, and then the discriminative model is sequentially updated by the incoming data and the selected label.

The fourth part is similar to the third part, but as shown in lines 11 and 12, a label is predicted by the discriminative model being retrained, and then the discriminative model is sequentially updated by the incoming data and the predicted label.

\begin{algorithm}[tb]
	\caption{Discriminative model reconstruction}
    \label{alg:reconst}
	\begin{algorithmic}[1]
		\Require discriminative model $model$, number of classes $C$, number of dimensions $D$, coordinates $cor$, numbers of samples in classes $num$, number to search initial coordinates $N_{search}$, number to update coordinates $N_{update}$, number to finish reconstruction $N$
	
\Function {Reconstruct\_Model}{$data$}
	\State $count \gets count +  1$
		\If{$count < N_{search}$}
	\State Init\_Coord($data$)
	\EndIf
		\If{$count < N_{update}$}
	\State Update\_Coord($data$)
	\EndIf
		\If{$count < N/2$}
		\State $label \gets \mathrm{argmin}_{c \in C}|cor[c] - data|$ 
		\State $model[label].\mathrm{training}(data)$
	\EndIf
		\If{$count < N$}
		\State $label \gets \mathrm{argmin}_{c \in C}model[c].\mathrm{predict}(data)$ 
		\State $model[label].\mathrm{training}(data)$
	\EndIf
	\If{$count = N$}
		\State $count \gets 0$
		\State \Return False
	\Else
		\State \Return True
	\EndIf
\EndFunction

\end{algorithmic}
\end{algorithm}

\begin{algorithm}[tb]
	\caption{Label coordinates initialization}
    \label{alg:initcoord}
	\begin{algorithmic}[1]
    \Require number of classes $C$, coordinates $cor$, number of dimensions $D$
\Function {Init\_Coord}{$data$}
	\State $label \gets -1$
	\State $min \gets \sum_{j=0}^{C-1}\sum_{k = j+1}^{C} \sum_{dim = 0}^{D} |cor[j][dim] - cor[k][dim]|$
	\For {$c \in C$}:
	\State {$tmp \gets cor[c]$}
	\State {$cor[c] \gets data$}
	\State $dist \gets \sum_{j=0}^{C-1}\sum_{k = j+1}^{C} \sum_{dim = 0}^{D} |cor[j][dim] - cor[k][dim]|$
	\State {$cor[c] \gets tmp$}
		\If{$min < dist$}
	\State {$label \gets c$}
	\State {$min \gets dist$}
	\EndIf
	\EndFor
		\If{$label \neq -1$}
	\State {$cor[label] \gets data$}
	\EndIf
\EndFunction

\end{algorithmic}
\end{algorithm}

\begin{algorithm}[tb]
	\caption{Label coordinates update}
    \label{alg:updatecoord}
	\begin{algorithmic}[1]
	\Require number of classes $C$, coordinates $cor$, numbers of samples in classes $num$
	
\Function {Update\_Coord}{$data$}
		\State $label \gets \mathrm{argmin}_{c \in C}|cor[c] - data|$
	\State $cor[label] \gets \dfrac{cor[label] \times num[label] + data}{num[label] + 1}$
	\State $num[label] \gets num[label] + 1$
\EndFunction

\end{algorithmic}
\end{algorithm}

\subsection{Threshold}\label{ssec:thr}
The threshold value $\theta_{drift}$ is used to detect a concept drift in line 17 of Algorithm \ref{alg:main}.
For each trained sample, a distance between the sample and the centroid of its predicted label is calculated and stored in $dist$ array.
In this paper, $\theta_{drift}$ is calculated based on the mean and standard deviation of $dist$ array, as shown below.
\begin{equation}
\label{eq:threshold}
	\begin{split} 
   \mu &= \frac{1}{N}\Sigma_{i \in N}{dist[i]}\\
   \theta_{drift} & = \mu + \sqrt{\frac{1}{N}\Sigma_{i \in N}{(dist[i] - \mu)^{2}}},
\end{split} 
\end{equation}
where $N$ is the number of trained samples and $dist[i]$ is a distance between the $i$-th sample and the centroid of a cluster the $i$-th sample belongs to.

\section{Evaluation Setup}\label{sec:env}
This section describes datasets, counterparts of the proposed method, and platforms for the evaluations.
\subsection{Datasets}
Two datasets used in the evaluations are described below.
\subsubsection{NSL-KDD Dataset}
NSL-KDD \cite{Datanet} is a famous dataset which can be used to evaluate network intrusion detection methods.
As data distribution of the dataset shifts from the training data to the test data, it can also be used for evaluating concept drift detection methods \cite{OASW}.
The original dataset contains a lot of samples with 23 labels. In this paper, we use selected samples labeled with ``normal'' and ``neptune'' which are the first and second largest labels. We further select 2522 and 22701 samples for the initial training and test, respectively.
A concept drift occurs at the 8333rd data point.
\subsubsection{Cooling Fan Dataset}\label{sssec:coolingfan}
The cooling fan dataset \cite{Datafan} contains vibration patterns of various cooling fans including normal and damaged ones measured by an industrial accelerometer PCB M607A11.
In the damaged fans, holes are made or an edge of one of blades is chipped. Since these damaged blades cause a radial unbalance of the cooling fan's mass, abnormal vibration patterns are generated.
These vibration patterns were measured at a silent environment and a noisy environment near a ventilation fan.
The vibration pattern is represented as a frequency spectrum ranging from 1Hz to 511Hz. Thus, the number of features is 511 in the cooling fan dataset.
Vibration patterns of a normal cooling fan without damages observed in a silent environment are used as a training dataset.
The following three patterns are used as test datasets.
\begin{enumerate}
\item The first dataset focuses on a sudden drift. Vibration patterns measured from a normal fan without damages in a silent environment are used as test data before a concept drift. Those of a damaged fan having holes in a blade in the same environment are used as test data after the concept drift. The concept drift occurs at the 120th data point.
\item The second dataset focuses on a gradual drift. Vibration patterns before a concept drift are the same as the first dataset, but those of a damaged fan with a chipped blade are used as test data after the concept drift. In the second dataset, both the patterns are mixed between the 120th data point and the 600th data point so that a gradual drift can be reproduced. 
\item The third dataset focuses on a reoccurring drift. Vibration patterns before and after a concept drift are the same as the second dataset, but those after the concept drift appear only between the 120th data point and the 170th data point. Those before the concept drift reoccur after the 170th data point so that a reoccurring drift can be examined. 
\end{enumerate}

\subsection{Evaluated Methods}\label{ssec:methods}
In this paper, the following five combinations are evaluated and compared as concept drift countermeasures.
\begin{enumerate}
    \item Detector: the proposed method, Discriminative model: OS-ELM
    \item Detector: none (no concept drift detection), Discriminative model: OS-ELM
    \item Detector: Quant Tree, Discriminative model: OS-ELM
    \item Detector: SPLL, Discriminative model: OS-ELM
    \item Detector: none, Discriminative model: ONLAD (OS-ELM with a forgetting mechanism)
\end{enumerate}
The first method is our proposal, and the second method is a baseline without concept drift detection. The first, third, and fourth methods are classified as the active detection approach, while the fifth method is the passive approach. 

In these methods, OS-ELM is used in the discriminative model.
More specifically, the same number of OS-ELM based autoencoder instances as the number of labels are used, as mentioned in Section \ref{ssec:model}.
The reason for using OS-ELM in the discriminative model is that it can be retrained on resource-limited edge devices \cite{Matutani}. Another reason is that we want to compare the proposed method with ONLAD that uses OS-ELM and a lightweight forgetting mechanism. 

Hyper-parameters for NSL-KDD dataset are as follows.
In the OS-ELM based discriminative model, the numbers of input and output layer nodes are 38 and that of the hidden layer nodes is 22.
In Quant Tree, the batch size is 480 and the number of histograms is 32. In SPLL, the batch size is 480.
In ONLAD, the numbers of input, hidden, and output layer nodes are the same as those of the OS-ELM model. The forgetting rate is 0.97.
Regarding the cooling fan dataset,
for the OS-ELM based discriminative model, the numbers of input and output layer nodes are 511 and that of the hidden layer nodes is 22.
In Quant Tree, the batch size is 235 and the number of histograms is 16. In SPLL, the batch size is 235.
In ONLAD, the numbers of input, hidden, and output layer nodes are the same as those of the OS-ELM model. The forgetting rate is 0.99.

\subsection{Evaluation Platforms}
The above-mentioned five methods are running on Raspberry Pi 4 Model B \cite{Raspi4}. In addition, only the proposed method is demonstrated on Raspberry Pi Pico \cite{RaspiPico}.
These methods are evaluated in terms of memory utilization in Raspberry Pi 4 Model B.
The execution time breakdown of the proposed method is further analyzed in Raspberry Pi Pico.
Table \ref{tb:raspi} shows the specifications of Raspberry Pi 4 Model B and Raspberry Pi Pico.
\begin{table}[htb]
    \centering
    \caption{Specifications of Raspberry Pi 4 Model B and Raspberry Pi Pico}
    \label{tb:raspi}
    \begin{tabular}{l|c|c} \hline\hline
	& Raspberry Pi 4 Model B & Raspberry Pi Pico \\ \hline
        OS  & Raspberry Pi OS & - \\
        CPU & ARM Cortex-A72, 1.5GHz & ARM Cortex-M0+, 133MHz \\
        RAM & 4GB & 264kB \\ \hline
    \end{tabular}
\end{table}

\section{Evaluation Results} \label{sec:eval}
This section shows evaluation results of the five methods listed in Section \ref{ssec:methods} in terms of the accuracy, delay to detect concept drifts, memory utilization, and execution time.

\subsection{Accuracy and Delay}
\begin{figure}[H]
        \centering
        \includegraphics[width=92mm]{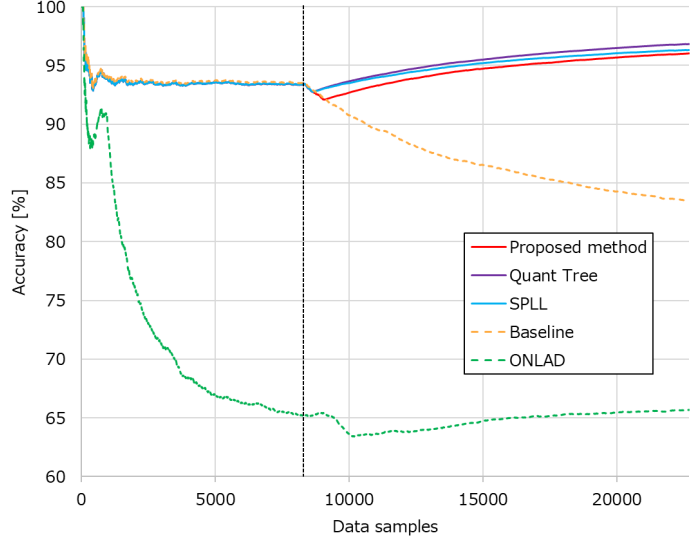}
        \caption{Accuracy changes on NSL-KDD dataset}
        \label{fig:accurancy1}
\end{figure}

The five methods are evaluated with NSL-KDD dataset.
Figure \ref{fig:accurancy1} shows evaluation results in terms of the accuracy of the discriminative model.
Table \ref{tb:performanceNSL} summarizes the accuracy and the delay to detect a concept drift. The delay means the number of samples needed to detect a concept drift after the concept drift actually happens.
A vertical bar at the 8333rd data point in Figure \ref{fig:accurancy1} shows a concept drift.

First, the results show that the parameter tuning of a forgetting rate of ONLAD is difficult.
Since a concept drift happens at the 8333rd data point, it is expected that accuracy of the ONLAD model should be constantly high before the concept drift.
However, the results show that the accuracy of the ONLAD model gradually decreases even before the concept drift happens.

The results also show that the proposed method can detect the concept drift as well as the batch-based Quant Tree and SPLL methods.
After the concept drift is detected, the accuracy of the proposed method becomes high compared to the baseline method that does not detect concept drifts.
As a result, the proposed method outperforms the baseline method without concept drift detection and ONLAD with a forgetting mechanism by 9.0\% and 26.8\%, respectively, while the accuracy is decreased by up to 3.8\% and 4.3\% compared to the batch-based SPLL and Quant Tree methods, respectively.

Please note that the proposed method needed more samples to detect the concept drift compared to the batch-based Quant Tree and SPLL methods.
The reason for the detection delay is that the distances between trained centroids and those after the concept drift are relatively small compared to the threshold value computed by Equation \ref{eq:threshold}. A manual tuning of the threshold value can shorten the detection delay.

\begin{table}[h!]
  \centering
  \caption{Accuracy (\%) and delay for detecting concept drift on NSL-KDD dataset}
  \label{tb:performanceNSL}
  \begin{tabular}{l|r|r}\hline \hline
	& Accuracy & Delay \\ \hline
	Quant Tree & 96.8 & 296 \\
	SPLL & 96.3 & 296 \\
	Baseline (no detector) & 83.5 & - \\
	ONLAD & 65.7 & - \\
	Proposed method (Window size = 100) & 96.0 & 843 \\
	Proposed method (Window size = 250) & 95.5 & 993 \\
	Proposed method (Window size = 1000) & 92.5 & 1263 \\ \hline
  \end{tabular}
\end{table}

\subsection{Window Size and Delay}
The proposed detection method is further analyzed in terms of the window size vs. delay to detect a concept drift after the concept drift actually happens at the 120th data point.
It is evaluated with the cooling fan dataset assuming the sudden, gradual, and reoccurring concept drifts.
Table \ref{tb:numdetect} summarizes the evaluation results.
As mentioned in Section \ref{sssec:coolingfan}, the cooling fan dataset was modified so that it can be used for the evaluations of the sudden drift, gradual drift, and reoccurring drift.
Using the modified dataset, the proposed method is analyzed in terms of the window size vs. detection delay for the three concept drift types.

\begin{enumerate}
\item The results show that a smaller window size can shorten the detection delay in the sudden drift case where an old data distribution before the concept drift does not appear again.
\item In the gradual drift case, an old data distribution is gradually replaced with a new data distribution. Since the old and new data distributions are mixed during the concept drift, a too small window size may detect short-term data changes as concept drifts. A larger window size than these short-term changes is better so that the concept drift does not oscillate between the old and new data distributions.
\item In our dataset for the reoccurring concept drift, a new data distribution appears between the 120th data point and the 170th data point, and the old data distribution reoccurs after the 170th data point. In Table \ref{tb:numdetect}, the new data distribution is detected as a concept drift when the window size is 10 and 50 while it is not detected when the window size is 150. If it is not expected to detect the new data distribution as a concept drift, a larger window size should be selected.
\end{enumerate}

Based on the above discussion, the window size should be determined by considering expected concept drift types and detection behaviors.
Using multiple detection models with different window sizes is our future work to address more complicated drift behaviors.

\begin{table}[h!]
    \centering
    \caption{Delay for detecting concept drift with different window sizes on cooling fan dataset}
    \label{tb:numdetect}
    \begin{tabular}{l|r|r|r}\hline \hline
	& Sudden & Gradual & Reoccurring \\ \hline
	Window size = 10 & 53 & 161 & 22 \\
	Window size = 50 & 60 & 157 & 62 \\
	Window size = 150 & 160 & 257 & - \\ \hline
    \end{tabular}
\end{table}

\subsection{Memory Utilization}
Table \ref{tb:resource} shows the evaluation results in terms of the memory utilization on Raspberry Pi 4 Model B. The cooling fan dataset is used for the memory size evaluation; in this case, the batch size of the Quant Tree and SPLL methods is 235 while it is one in the proposed method.

The results show that the proposed method uses much less memory size compared to the batch-based Quant Tree and SPLL methods.
Specifically, the proposed method decreases the memory utilization by up to 96.4\% and 88.9\% compared to SPLL and Quant Tree, respectively.
This is because in the batch-based concept drift detection methods, data samples are stored in the device memory to detect concept drifts, while the proposed method processes data samples one by one and detects concept drifts sequentially; thus, the proposed method does not store past samples in the device memory.

Please note that since RAM size of Raspberry Pi Pico is only 264kB as shown in Table \ref{tb:raspi}, the batch-based Quant Tree and SPLL methods cannot operate on Raspberry Pi Pico.
In Section \ref{sec:speed}, only the proposed method shows the execution time on Raspberry Pi Pico in addition to that on Raspberry Pi 4 Model B.

\begin{table}[h!]
    \centering
    \caption{Memory utilization (kB)}
    \label{tb:resource}
    \begin{tabular}{l|r}\hline \hline
	& Memory size \\ \hline
	Quant Tree & 619 \\ 
	SPLL & 1933 \\
	Proposed method & 69 \\ \hline
    \end{tabular}
\end{table}

\subsection{Execution Time}\label{sec:speed}
Tables \ref{tb:speed4b} and \ref{tb:speedpico} show evaluation results in terms of the execution times on Raspberry Pi 4 Model B and Raspberry Pi Pico, respectively. The same cooling fan dataset is used for this evaluation.

Table \ref{tb:speed4b} shows the execution time to process the cooling fan dataset that contains 700 samples in total.
As shown, the execution time of the proposed method is slightly less than that of Quant Tree and it is much less than that of SPLL.
Specifically, the proposed method decreases the execution time by up to 83.8\% and 1.3\% compared to SPLL and Quant Tree, respectively.
Since SPLL executes k-means clustering, the execution time of SPLL is increased compared to the others.
Although the proposed method increases the execution time by 42.9\% compared to the baseline method without concept drift detection, it significantly improves the accuracy compared to the baseline as shown in Figure \ref{fig:accurancy1}.
Thus, the proposed method is an attractive option when concept drifts are expected in target applications.

Table \ref{tb:speedpico} further analyzes the execution time breakdown for a single sample by the proposed method on Raspberry Pi Pico.
In the table, ``Label prediction'' and ``Distance computation'' are corresponding to lines 6 and 14 in Algorithm \ref{alg:main}, respectively.
``Model retraining without label prediction'' is done in lines 8 and 9, while
``Model retraining with label prediction'' is done in lines 11 and 12 in Algorithm \ref{alg:reconst}.
``Label coordinates initialization'' is Init\_Coord() function in Algorithm \ref{alg:initcoord}, and
``Label coordinates update'' is Update\_Coord() function in Algorithm \ref{alg:updatecoord}.
The results show that the additional computation time for the concept drift detection is less than the label prediction time of the discriminative model.
Please note that the latency is within a few hundred milliseconds even in such a low-end edge device.

\begin{table}[h!]
    \centering
    \caption{Execution time (sec) for 700 samples on Raspberry Pi 4 Model B}
    \label{tb:speed4b}
    \begin{tabular}{l|r}\hline \hline
	& Execution time \\ \hline
	Quant Tree & 1.52 \\
	SPLL & 9.28 \\
	Baseline (no detector) & 1.05 \\
	Proposed method & 1.50 \\ \hline
    \end{tabular}
\end{table}

\begin{table}[h!]
    \centering
    \caption{Execution time breakdown (msec) for 1 sample by proposed method on Raspberry Pi Pico}
    \label{tb:speedpico}
    \begin{tabular}{l|r}\hline \hline
	& Execution time \\ \hline
	Label prediction & 148.87 \\
	Distance computation & 10.58 \\
	Model retraining without label prediction & 25.42 \\
	Model retraining with label prediction & 166.65 \\
	Label coordinates initialization & 25.59 \\
	Label coordinates update & 6.05 \\ \hline
    \end{tabular}
\end{table}

\section{Summary} \label{sec:conc}
In edge AI systems, data distributions of trained dataset and deployed environment may differ due to noise and environmental changes over time.
This gap degrades the performance of edge AI systems and may introduce system failures. To address this gap, retraining of neural network models triggered by concept drift detection is a practical approach.
As practical concept drift detection, error-rate based detection methods and distribution-based detection methods have been used.
However, since such a batch-based processing and use of labeled teacher dataset are not suited for resource-limited edge devices, in this paper we proposed a fully sequential concept drift detection method in cooperation with the on-device sequential learning technique of neural networks.
Both the neural network retraining and the proposed concept drift detection are done only by sequential computation to reduce computation cost and memory utilization.

Evaluation results of the proposed approach showed that while the accuracy is decreased by 3.8\%-4.3\% compared to existing batch-based detection methods, it decreases the memory size by 88.9\%-96.4\% and the execution time by 1.3\%-83.8\%. Thanks to these significant decreases on the memory size and computation cost, the combination of the neural network retraining and the proposed concept drift detection method was demonstrated on Raspberry Pi Pico that has 264kB memory.

A possible extension of this work is a combination of multiple detection models with different window sizes to address more complicated concept drift behaviors.
We are also planning to evaluate the proposed method with more concept drift datasets to emphasize the benefits.


\begin{thebibliography}{10}

\bibitem{Zhou19}
Zhi Zhou, Xu~Chen, En~Li, Liekang Zeng, Ke~Luo, and Junshan Zhang.
\newblock {Edge Intelligence: Paving the Last Mile of Artificial Intelligence
  With Edge Computing}.
\newblock {\em Proceedings of the IEEE}, 107(8):1738--1762, 2019.

\bibitem{Matutani}
Hiroki Matsutani, Mineto Tsukada, and Masaaki Kondo.
\newblock {On-Device Learning: A Neural Network Based Field-Trainable Edge AI}.
\newblock arXiv Preprint 2203.01077, 2022.

\bibitem{OASW}
Li~Yang and Abdallah Shami.
\newblock {A Lightweight Concept Drift Detection and Adaptation Framework for
  IoT Data Streams}.
\newblock {\em IEEE Internet of Things Magazine}, 4(2):96--101, 2021.

\bibitem{Survey}
Jie Lu, Anjin Liu, Fan Dong, Feng Gu, Joao Gama, and Guangquan Zhang.
\newblock {Learning under Concept Drift: A Review}.
\newblock {\em {IEEE} Transactions on Knowledge and Data Engineering},
  31(12):2346--2363, 2019.

\bibitem{NOVELOSELM}
Zhe Yang, Sameer Al-Dahidi, Piero Baraldi, Enrico Zio, and Lorenzo Montelatici.
\newblock A novel concept drift detection method for incremental learning in
  nonstationary environments.
\newblock {\em IEEE Transactions on Neural Networks and Learning Systems},
  31(1):309--320, 2020.

\bibitem{OSELM}
{Nan-ying} Liang, {Guang-bin} Huang, P.~Saratchandran, and N.~Sundararajan.
\newblock {A Fast and Accurate Online Sequential Learning Algorithm for
  Feedforward Networks}.
\newblock {\em IEEE Transactions on Neural Networks}, 17(6):1411--1423, 2006.

\bibitem{OSUAD}
Mineto Tsukada, Masaaki Kondo, and Hiroki Matsutani.
\newblock {A Neural Network-Based On-device Learning Anomaly Detector for Edge
  Devices}.
\newblock {\em IEEE Transactions on Computers}, 69(7):1027--1044, 2020.

\bibitem{Survey2}
Jo{\~a}o Gama, Indre Zliobaite, Albert Bifet, Mykola Pechenizkiy, and
  A.~Bouchachia.
\newblock {A Survey on Concept Drift Adaptation}.
\newblock {\em ACM Computing Surveys}, 46(4):1--37, 2014.

\bibitem{DDM}
Joao Gama, Pedro Medas, Gladys Castillo, and Pedro Rodrigues.
\newblock {Learning with Drift Detection}.
\newblock In {\em Proceedings of the Brazilian Symposium on Artificial
  Intelligence}, pages 286--295, 2004.

\bibitem{ADWIN}
Albert Bifet and Ricard Gavalda.
\newblock {Learning from Time-Changing Data with Adaptive Windowing}.
\newblock In {\em Proceedings of the SIAM International Conference on Data
  Mining}, pages 443--448, 2007.

\bibitem{QuantTree}
Giacomo Boracchi, Diego Carrera, Cristiano Cervellera, and Danilo Macci{\`o}.
\newblock {Quant Tree: Histograms for Change Detection in Multivariate Data
  Streams}.
\newblock In {\em Proceedings of the International Conference on Machine
  Learning}, pages 639--648, 2018.

\bibitem{SPLL}
Ludmila~I. Kuncheva.
\newblock {Change Detection in Streaming Multivariate Data Using Likelihood
  Detectors}.
\newblock {\em IEEE Transactions on Knowledge and Data Engineering},
  25(5):1175--1180, 2013.

\bibitem{Hinton06}
G.~Hinton and R.~Salakhutdinov.
\newblock {Reducing the Dimensionality of Data with Neural Networks}.
\newblock {\em Science}, 313(5786):504--507, 2006.

\bibitem{Arthur07}
David Arthur and Sergei Vassilvitskii.
\newblock {k-means++: The Advantages of Careful Seeding}.
\newblock In {\em Proceedings of the ACM-SIAM Symposium on Discrete
  Algorithms}, pages 1027--1035, 2007.

\bibitem{Datanet}
{NSL-KDD Dataset Research Candian Institute for Cybersecurity}.
\newblock \url{https://www.unb.ca/cic/datasets/nsl.html}.

\bibitem{Datafan}
{A cooling fan dataset containing vibration patterns of various fans at silent
  and noisy environments}.
\newblock \url{https://github.com/matutani/cooling-fan}.

\bibitem{Raspi4}
{Raspberry Pi 4}.
\newblock \url{https://www.raspberrypi.com/products/raspberry-pi-4-model-b/}.

\bibitem{RaspiPico}
{Raspberry Pi Pico}.
\newblock \url{https://www.raspberrypi.com/products/raspberry-pi-pico/}.

\end{thebibliography}

\end{document}